\pdfoutput=1
\documentclass[11pt]{article}

\usepackage[]{EMNLP2023}

\usepackage{times}
\usepackage{latexsym}
\usepackage{graphicx}
\usepackage{epsfig} 
\usepackage{booktabs}
\usepackage{array}
\usepackage{makecell}
\usepackage{tabularx}
\usepackage[utf8]{inputenc}
\usepackage{multirow}
\usepackage{float}
\usepackage{caption}
\usepackage{geometry}
\usepackage{icomma}
\usepackage{subcaption}
\usepackage[T1]{fontenc}
\usepackage[utf8]{inputenc}

\usepackage{microtype}

\usepackage{inconsolata}
\usepackage{enumitem}

\newcommand{\ie}{\textit{i}.\textit{e}.}

\newcommand{\para}[1]{\noindent \textbf{#1\quad}}
\newcommand{\ours}{PreWoMe}

\title{PreWoMe: Exploiting Presuppositions as Working Memory \\for Long Form Question Answering}

\author{Wookje Han \\
  Columbia University \\
  \texttt{wookje.han@columbia.edu} \\\And
  Jinsol Park \\
  Carnegie Mellon University\\
  \texttt{jinsolp@cs.cmu.edu} \\\And
  Kyungjae Lee\thanks{~~Corresponding author} \\ 
  LG AI Research \\
  \texttt{kyungjae.lee@lgresearch.ai}
}    

\begin{document}
\maketitle
\begin{abstract}
Information-seeking questions in long-form question answering (LFQA) often prove \textit{misleading} due to \textit{ambiguity} or \textit{false presupposition} in the question.
While many existing approaches handle misleading questions, they are tailored to limited questions, which are insufficient in a real-world setting with unpredictable input characteristics.
In this work, we propose \ours{}, a unified approach capable of handling any type of information-seeking question. 
The key idea of \ours{} involves extracting presuppositions in the question and exploiting them as working memory to generate feedback and action about the question. 
Our experiment shows that \ours{} is effective not only in tackling misleading questions but also in handling normal ones, thereby demonstrating the effectiveness of leveraging presuppositions, feedback, and action for real-world QA settings.
\end{abstract}

\section{Introduction}
\label{sec:intro}
Answering information-seeking long-form questions has recently shown significant progress \citep{eli5, hurdles, evalLFQA}.
However, users' questions in the real-world may often \textit{mislead} language models (LMs) to output misinformation.
For example, users ask questions with \textbf{false presuppositions} (FP) \citep{qaqa, crepe} which can induce hallucinations if LMs believe the presuppositions are true \citep{hallucination}. 
Also, users often ask questions that do not have a single and clear answer, \ie, an \textbf{ambiguous} question, which is difficult for LMs to identify~\cite{ambigqa,asqa}.

\begin{figure}[t]
  \centering
  \includegraphics[width=\linewidth]{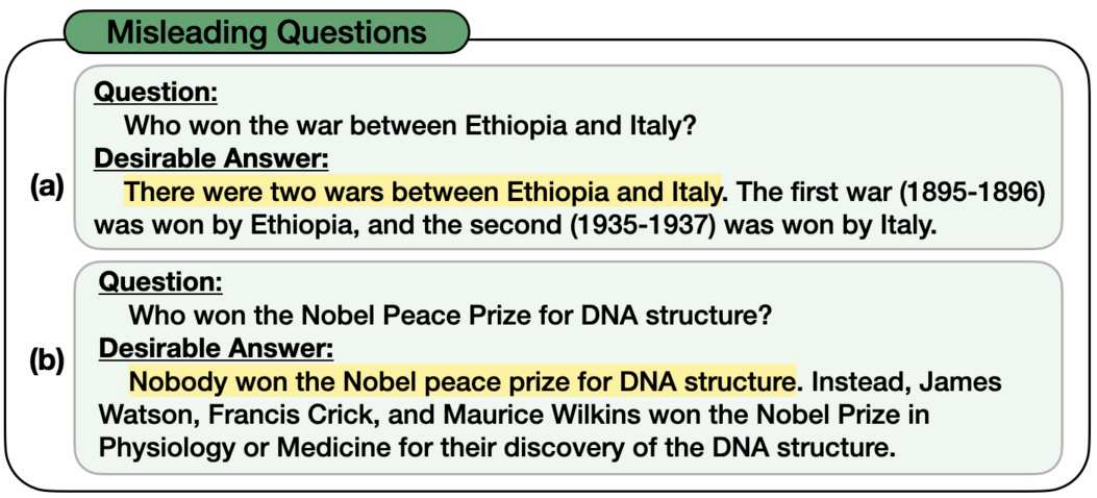}
  \caption{Two types of misleading questions: (a) having ambiguity, and (b) false presupposition. Highlights show essential cues to avoid misleading.}
 \label{fig:imperfect_example}
\end{figure}

\autoref{fig:imperfect_example} shows examples of two types of misleading questions: (a) an ambiguous question that requires considering prior knowledge (``two different wars between Italy and Ethiopia'') to answer properly, and (b) a question with FP (``somebody won the Nobel peace prize for DNA structure''), which should be corrected before answering the question \cite{crepe,qaqa}.

Desirable answers to misleading questions require discerning and resolving misleadings.
A previous work by \citet{queryRefinementPrompt} focuses on ambiguous questions, by introducing query refinement prompts that encourage LMs to consider multiple facets of the question.
Another work \cite{linguistlightbulb} tackles questions containing FP, by extracting and verifying presuppositions.
While such works assume only one specific type of misleading factor, in a real scenario, the users ask various types of questions, which can be misleading or normal (\ie, non-misleading).\footnote{Throughout this paper, we use the term \textit{misleading questions} for questions that are ambiguous or have FP.}
This makes prior single-type-tailored approaches hard to be deployed in real-world QA settings.

In this paper, we emphasize the role of presuppositions in handling \textit{any type} of information-seeking questions in a \textit{unified} way.
Presuppositions are crucial building blocks of questions and play an essential role in understanding the meaning of the question \citep{presup_all, presup_question}. 
This suggests that exploiting presuppositions is promising in handling misleading questions, while generally applicable to any type of question.
Based on this, we propose a new approach, \textbf{\ours{}} (\textbf{Pre}supposition as \textbf{Wo}rking \textbf{Me}mory), that handles \textit{any type} of information-seeking question in a \textit{unified} way without any parameter updates.

The main idea of \ours{} is to extract presuppositions and use them as working memory to generate analysis and directions for answering the question.
Our contributions can be summarized as follows:
\vspace{-0.6em}
\begin{itemize}[leftmargin=0.4cm]
\itemsep-0.3em
   \item We analyze the performance of recent large language models (LLMs), GPT-4 and GPT-3.5, on misleading questions. To the best of our knowledge, we are the first to explore the performance of LLMs this large on misleading questions.
   \item We propose \ours{}, a new approach designed for real-world QA settings that 1) is capable of handling any type of information-seeking question, and 2) does not entail any parameter updates, making it easily adaptable to LLMs.
   \item We propose using presuppositions as a working memory when answering information-seeking questions.
\end{itemize}

\section{Background}
\label{Background}

\para{Presupposition} Presuppositions are conditions that are believed to be true by speakers in a discourse \citep{crepe}. 
Thus, in order for an utterance to be \textit{appropriate}, the presuppositions should be true.
For example, if someone said \textit{I care for my sister}, then it can be assumed that the speaker has a sister.
If not, the utterance would be \textit{inappropriate}.
In this paper, we suggest using presuppositions to solve an information-seeking QA task.

\para{Chain-of-thought Prompting (CoT)} 
\citet{cot} explores how thinking with intermediate steps greatly improves LM's performance. 
Chain-of-thought Prompting effectively enhances the reasoning abilities of LMs, particularly in mathematical and logical reasoning tasks. 
This technique involves guiding the model through a series of interconnected thought processes. 
Inspired by this concept, we propose a novel approach that leverages presupposition, feedback, and action, serving as intermediate steps, which boosts the robustness of LMs in handling open-ended questions.

\para{Working Memory}
Working memory \citep{working_memory} is the immediate information buffer that is accessed while performing conscious tasks \citep{llmWorkMem}.
We make presuppositions function as working memory by extracting and feeding them into the model along with the question as intermediate steps.

\para{Self-correcting Approaches} 
While LLMs occasionally struggle to generate accurate answers without hallucination, recent works suggest that LLMs can enhance accuracy by \textit{self-correcting} their responses through iterative prompting~\citep{madaan2023self,reflex, selfcritique,press2022measuring,selfverif}.
Such approaches are based on the idea that verifying correctness is easier than directly generating an accurate answer.
Inspired by this idea, we hypothesize that 1) LLMs are capable of generating and verifying presuppositions implicit in a given question, and 2) guiding LLMs to generate answers \textit{after} the verification will yield better results than generating answers directly.

\begin{table*}[t]
\scriptsize	
\centering
\renewcommand{\arraystretch}{0.6}
\begin{center}
\begin{tabularx}{\textwidth}{c|X|X}
\toprule
\textbf{Type} & \textbf{(a) Ambiguous} & \textbf{(b) False Presupposition}  \\
\midrule
% \midrule
Question & Who won the war between Ethiopia and Italy?
 & Who is the only Indian to win the Oscar for music?
 \\
\midrule
Presuppositions & 
1) There was a war between ethiopia and italy. \par 
2) Some country won the war. \par
A) There is a clear and single answer to the question.& 
1) There is a category for music in the Oscars. \par
2) There is only one Indian who won the Oscar for music. \par
A) There is a clear and single answer to the question.\\
\midrule
Feedback & 
There were two wars between Italy and Ethiopia, with different winners. Therefore, the question \textbf{contains a false presupposition that there is a clear and single answer to the question.} &
The question \textbf{contains a false presupposition that there is only one Indian who won the Oscar for music.} \\
\midrule
Action & 
Your answer should include the winner of the war between Ethiopia and Italy for \textbf{each war respectively in detail}.&
\textbf{Correct the false presupposition} that there is only one Oscar winner Indian and respond based on the corrected presupposition.\\
\bottomrule
\end{tabularx}
\end{center}
\caption{Examples of misleading questions and their intermediate steps. Boldface in feedback and action are parts that distinguish different types of questions.}
\label{table:example}
\end{table*}

\begin{figure}[t]
  \centering
  \includegraphics[width=92mm]{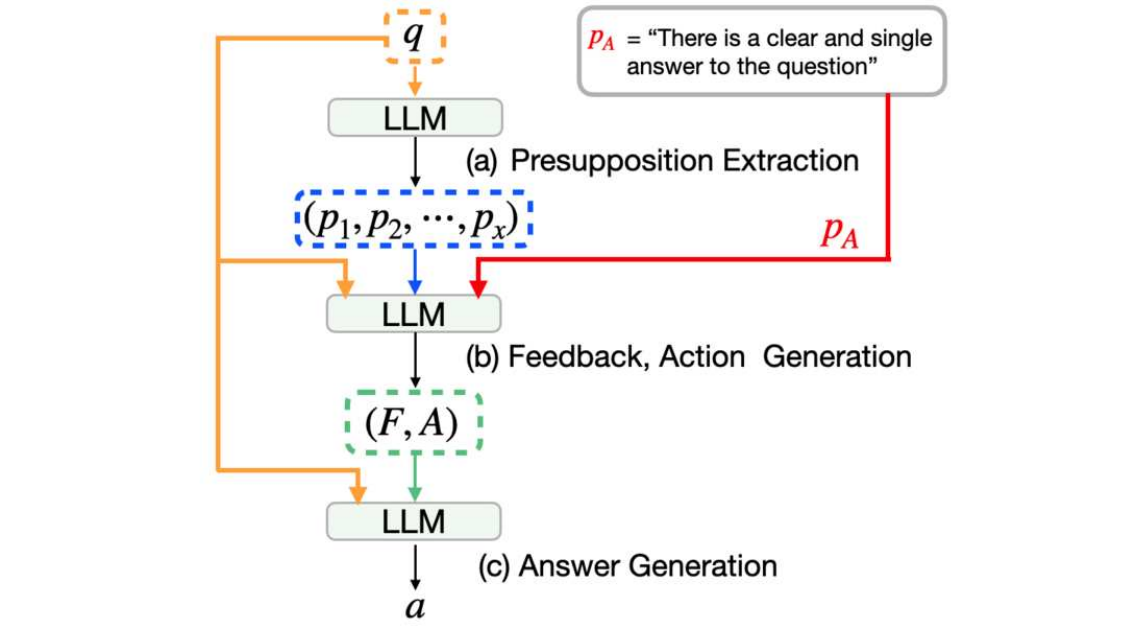}
  \caption{Overall pipeline of \ours{}. $q$ is the given question, ($p_1, p_2, ..., p_x$) are the model-generated presuppositions. $(F, A)$ is feedback and action.}
 \label{fig:method}
\end{figure}

\section{\ours{}}
We introduce \ours{}, a new approach that answers any type of long-form information-seeking question in a unified way.
\ours{} follows three intermediate steps, as illustrated in \autoref{fig:method}.

\subsection{Presupposition Extraction} \label{Presupposition Extraction}
To exploit presuppositions as working memory, \ours{} first generates presuppositions contained in the question $q$.
Motivated by the few-shot learning capability of LLMs ~\citep{gpt3}, we feed $q$ with $N$ (question, presupposition) pairs as few-shot examples within a designed prompt (See Appendix \ref{sec:appendix2-designed prompt}).
The LLM generates multiple presuppositions ($p_1, p_2,\cdots, p_x$) in $q$ using this input (\autoref{fig:method}(a)).

In addition, we append a manual presupposition $p_A$: \textit{``There is a clear and single answer to the question''}, as shown in \autoref{fig:method}, which is a common presupposition of all normal questions. If $p_A$ is recognized as False in the next step, the question is classified as \textit{misleading} due to its ambiguity.

\subsection{Feedback and Action Generation} \label{Feedback and Action}

Motivated by \citet{react} that explores the effectiveness of reasoning processes and task-specific actions, we make LLMs to generate \textit{feedback} and \textit{action} for $q$ (\autoref{fig:method}(b)) by exploiting extracted presuppositions as working memory.
We concatenate $q$ with extracted presuppositions and $p_A$, and feed it as part of an input to the LLM.
We use $N$ (question, presupposition, feedback-action) triplets as few-shot examples within a designed prompt (See \autoref{sec:appendix2-designed prompt}).

\autoref{table:example} shows examples of the generated feedback and action for two types of misleading questions.
The \textit{feedback} is the description of whether the given input contains FP or not.
It can be seen that both types of misleading questions can be handled in a unified way -- by indicating which presupposition is false.
If the LLM determines that $p_A$ is false, the feedback provides the hint that question $q$ is ambiguous.
If other presuppositions are determined to be false, then the corresponding feedback states that question $q$ contains FP.
For normal questions, the feedback informs that all presuppositions are valid (See \autoref{sec:appendix1-normalQ example}).
The feedbacks can be viewed as a classifier that determines the type of input question.

The \textit{action} is generated coherently with the feedback, and serves as an instruction to answering the question.
\autoref{table:example} shows that for ambiguous questions, the action asks to specify the answer in detail for multiple facets, and for questions with FP, the action asks to correct the FP.

\subsection{Answer Generation}\label{Generation Answer}
When generating the final answer $a$, the feedback and action $(F, A)$ from the previous step act as a guideline that the language model can refer to.
We concatenate $q$ and $(F, A)$ and feed it into the model with $N$ pairs of (question, feedback-action, answer) as few-shot examples within a designed prompt (See \autoref{sec:appendix2-designed prompt}).

\begin{table*}[t]
\footnotesize
\centering
\renewcommand{\arraystretch}{0.8}
\setlength{\tabcolsep}{7.5pt}
\begin{center}
\begin{tabular}{c|l|ccc|ccc|ccc}
\toprule

&
\multirow{ 2}{*}{Method} &
\multicolumn{3}{c|}{(a) ASQA} & 
\multicolumn{3}{c|}{(b) (QA)$^2$} & 
\multicolumn{3}{c}{(c) BioASQ} \\

% \hline
&& R-L & D-F1 & DR &
R-1 & R-L & B-RT &
R-1 & R-L & B-RT \\

\midrule
\midrule
GPT-4&
Vanilla &
43.66 & 34.71 & 38.93 &
24.77 & 22.77 & 0.42 &
18.34 & 20.48 & 0.45 \\

&CoT &
49.16 & 34.59 & 41.24 &
26.12 & 24.57 & 0.43 &
20.16 & \textbf{22.42} & 0.46 \\

&Query Refinement &
49.57 & \textbf{35.32} & 41.84 &
-- & -- & -- &
-- & -- & -- \\

&Step-by-Step w. TD &
-- & -- & -- &
26.31 & 24.7 & 0.44 &
-- & -- & -- \\

&\ours{} (Ours) &
\textbf{51.04} & 35.11 & \textbf{42.34} &
\textbf{30.04} & \textbf{28.30} & \textbf{0.45} &
19.85 & 22.12 & \textbf{0.48} \\

& ~ - ~ w/o. Presup &
42.68 & 33.41 & 37.76 &
24.67 & 22.89 & 0.42 &
17.96 & 20.23 & 0.46 \\

& ~ - ~ w/o. $(F,A)$ &
43.20 & 34.31 & 38.50 &
25.35 & 23.27 & 0.43 &
\textbf{20.87} & 22.19 & 0.46 \\

\midrule

GPT-3.5&
Vanilla &
44.83 & 30.52 & 36.99 &
24.37 & 22.49 & 0.42 &
19.65 & 22.43 & 0.46 \\

&\ours &
39.58 & 28.45 & 33.55 &
26.54 & 24.80 & 0.43 &
17.72 & 20.35 & 0.46 \\

& ~ - ~  w. GPT-4 $(F,A)$ &
\textbf{50.66} & \textbf{31.63} & \textbf{40.03} &
\textbf{28.54} & \textbf{26.56} & \textbf{0.44} &
\textbf{19.90} & \textbf{22.50} & \textbf{0.49} \\
\bottomrule
\end{tabular}
\end{center}

    \caption{Results on GPT-4 and GPT-3.5. ``CoT'' is \citet{ZCOT}. ``Query Refinement'' is proposed in \citet{queryRefinementPrompt}. ``Step-by-Step w. TD'' is method using task decomposition (TD) proposed in \citet{qaqa}.``w/o. Presup'' is generating feedback-action \textit{without} using presuppositions. ``w/o. $(F,A)$'' is using only presupposition \textit{without} feedback and action. ``w. GPT-4 $(F,A)$'' is using feedback-action from GPT-4 to generate answers on GPT-3.5.}
  
  \label{table:main_exp}
\end{table*}

\section{Experiments and Results}
\label{Experiments}

\para{Datasets} We evaluate \ours{} on three distinct types of questions.
For misleading questions with ambiguity, we utilize ASQA \citep{asqa}, a LFQA dataset specifically curated for answering ambiguous questions.
For misleading questions with False Presupposition (FP), we use the subset of (QA)$^2$ \citep{qaqa} that are labeled as harboring FP.
For normal questions, we use BioASQ \citep{bioasq}, considering that questions in BioASQ are unlikely to be ambiguous or have FP as biomedical experts have carefully curated the questions.
Detailed statistics for each dataset are in Appendix \ref{subsec:appendix3-datasetstat}.

\para{Evaulation} For ASQA, we report Rouge-L \citep{rouge}, D-F1 (Disambiguated F1), and DR (Disambiguation-ROUGE) following ~\citet{asqa}.
Detail of each metric is in Appendix \ref{subsec:appendix-metrics}.
However, rather than using the f-measure, we propose using \textit{recall-measure} of Rouge-L instead.
This is because we notice that ASQA does not provide answers to all aspects of ambiguity, and thus using the f-measure against ASQA cannot fully reflect the comprehensiveness of the generated answer, which is also discussed in \citet{queryRefinementPrompt} (See Appendix \ref{subsec:appendix4-example_asqa}).
For (QA)$^2$ and BioASQ, we report f1-measure of Rouge-1, Rouge-L, and BleuRT \citep{BLEURT}.

\para{Models}
We evaluate \ours{} on two LLMs -- GPT-3.5 (\texttt{gpt-3.5-turbo}) and GPT-4 (\texttt{gpt-4}) \citep{gpt4} -- provided by OpenAI with default hyperparameters and temperature set to 0.

\subsection{Main Experiment} \label{main_exp}
We compare \ours{} with the vanilla LLM, which involves providing only a question without any feedback or action $(F,A)$.
We use six question-answer pairs as few-shot examples ($N=6$) for both \ours \ and vanilla LLM.
We also compare \ours\ with Chain-of-Thought Prompting (``CoT''), following the prompt format of \citet{ZCOT}.
As additional baselines, we consider two approaches that each target a single type of misleading question:
\citet{queryRefinementPrompt} proposed using refined prompts to make LLMs explicitly consider multifaceted aspects of questions to handle ambiguous questions (``Query Refinement'' in \autoref{table:main_exp}). 
\citet{qaqa} proposed a prompt that combine CoT \citep{ZCOT} with Task Decomposition (TD) prompting \citep{taskdecomp} to handle questions with FP (``Step-by-Step w. TD'' in \autoref{table:main_exp}).
To validate the capability of \ours{} on misleading questions, we compare \ours \ against these two additional baselines for each corresponding target dataset.

As shown in \autoref{table:main_exp}, we found that for GPT-4, our approach performs well on misleading questions as well as normal questions, outperforming the vanilla LLM across all datasets. Additionally, \ours{} outperforms CoT on misleading questions (ASQA and (QA)$^2$) and yields results comparable to those of CoT for BioASQ. It is noteworthy that CoT itself demonstrates better performance than the vanilla LLM, supporting that utilizing intermediate steps improves QA performance in LLMs.
In addition, \ours{} outperforms both of the additional baselines.
Especially considering that Query Refinement was optimized for ASQA and Step-by-Step w. TD was designed for (QA)$^2$, these findings suggest that our approach is versatile in handling arbitrary types of misleading questions.

For GPT-3.5, we found that the vanilla LLM shows better performance on some datasets than \ours{}.
Through manual analysis, we observed that GPT-3.5 lacks the ability to generate high-quality feedbacks and actions, compared to GPT-4 (See \autoref{sec:appendix8-FGquality}). 
Thus we conducted experiments
where we replaced feedback and action $(F,A)$ with those generated by GPT-4 and generated the answer on GPT-3.5.
The results outperformed not only the vanilla GPT-3.5, but also the vanilla GPT-4, demonstrating that 1) misleading $(F,A)$ can induce noise, 2) high-quality $(F,A)$ is transferable between different models, and 3) can positively impact LLMs to the extent where GPT-3.5 outperforms GPT-4.

For further validation on different N-shot ($N=4, 8$), we conducted additional experiments, which can be found in \autoref{sec:appendix7-diffshot}.

\subsection{Human Evaluation}
\begin{figure}[t]
  \centering
  \includegraphics[width=\linewidth]{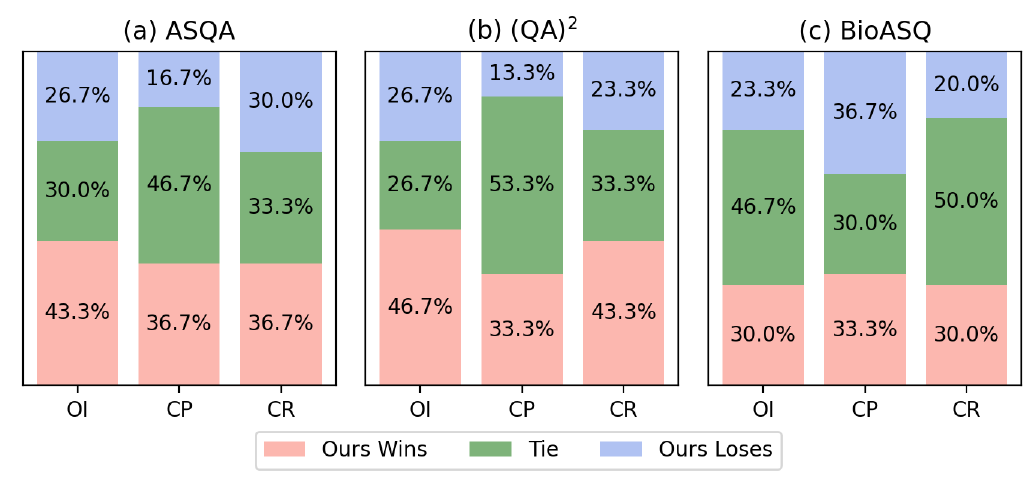}
  \caption{Human evaluation results for each dataset. We compare \ours with vanilla LLM (GPT-4) over 90 questions in total.}
 \label{fig:human_eval}
\end{figure}

For human evaluation, we randomly sample a total of 90 questions from three datasets and provided two generated answers, one from \ours{} and one from the vanilla LLM, with the golden answer to three human evaluators.
For each criterion -- overall impression (\textit{OI}), information completeness (\textit{CP}), and correctness (\textit{CR}) -- we ask human evaluators to choose a better answer between two generated answers or mark them as a tie, inspired by \citet{asqa}.
Then, we adopt a majority voting policy for each question based on the votes given by each human evaluator.
\autoref{fig:human_eval} shows that \ours{} outperforms the vanilla LLM for OI. 
The notable difference between \ours{} and vanilla LLM in CP for ASQA and CR for (QA)$^2$ suggests that \ours{} is capable of effectively covering more aspects of ambiguity and correcting FP when facing misleading questions.
The prompt used for human evaluation is provided in \autoref{sec:appendix8-human_eval_prompt}.

\subsection{Ablation Studies} \label{subsec:ablation}
For ablation studies, we first conduct analysis on the LLM's capability of handling each intermediate step, then verify the importance of each intermediate step within \ours{}.

\para{Presupposition Analysis} 
To analyze the LLM's capability of generating the correct presuppositions, 
we randomly sample 150 questions and manually annotate whether the corresponding 395 generated presuppositions were indeed assumed by the question.
The accuracy of presuppositions was extremely high (391 correct out of 395 total $\approx$ 99\%), demonstrating GPT-4’s capability of extracting accurate presuppositions.

\begin{table}[t]
\scriptsize
\centering

\renewcommand{\arraystretch}{0.7}
\setlength{\tabcolsep}{4pt}
\begin{center}
\begin{tabular}{c|c|ccc|cc}
\toprule
\multicolumn{2}{c|}{}&\multicolumn{3}{c|}{\textit{Prediction}} & \multicolumn{2}{c}{\textit{Performance}}\\
\multicolumn{2}{c|}{}&Normal & Ambig. & FP & Correct & Incorrect\\ 
\midrule

  \parbox[t]{1mm}{\multirow{6}{*}{\rotatebox[origin=c]{90}{~~ Golden}}} &Normal & \textbf{74.9\%} & 18.5\% & 6.6\%  & 23.19 / 21.26 & 19.03 / 18.25\\
\cmidrule{2-5}
&Ambig. & 54.6\% & \textbf{35.9\%} & 9.5\% & 39.70 / 34.04 & 43.65 / 41.51\\
\cmidrule{2-5}
&FP & 11.6\% & 14.0\% & \textbf{74.4\%} & 30.45 / 22.38 & 21.93 / 24.02\\
\bottomrule
\end{tabular}
\end{center}

  \caption{\textit{Prediction} shows distribution of predicted question type of \ours{} using generated feedback. \textit{Performance} indicates result of (\ours{} / vanilla LLM) with correct and incorrect predictions.}
  \label{table:classification}
\end{table}

\para{Feedback Analysis}
As mentioned in \autoref{Feedback and Action}, feedback works as a classifier that determines the type of the given question.
We extract the question type predicted by the feedback using a rule-based approach and report the distribution in \autoref{table:classification} (\textit{Prediction)}.
Feedback effectively detects normal questions and questions with FP (74.9\% and 74.4\%). 
However, ambiguous questions have low detect ratio, consistent with \citet{ambigqa}.
This indicates that there is more room to optimize for detecting ambiguity in the question, which we leave for future work.

We also report the final performance of \ours{} and vanilla LLM when given correctly and incorrectly classifying feedback (\autoref{table:classification} \textit{Performace}). 
Specifically, we report D-F1 for ASQA and Rouge-L for BIOASQ and (QA)$^2$ of \ours{} and vanilla GPT-4.
It can be seen that regardless of the question type, performance gains increase when feedback correctly classifies the question type.

\para{Importance of Presuppositions}
Do presuppositions really contribute to generating better feedback and action?
We explore this question by omitting presupposition extraction step and generating feedback and action with GPT-4 just based on the question.
As shown in \autoref{table:main_exp} (w/o. Presup), \ours{} without presuppositions extraction shows poor performance, highlighting the role of presuppositions as working memory.
One interesting observation is that the performance of \ours{} without presupposition is worse than the vanilla LLM, indicating that misleading $(F,A)$ becomes a noise, consistent with findings in \autoref{main_exp}.

\para{Importance of Feedback, Action}
Similarly, we also explore the importance of Feedback and Action. 
We omit the feedback and action generation step and just feed the questions with the presuppositions that are generated in the previous step into GPT-4.
\autoref{table:main_exp} (w/o. $(F,A)$) shows that using presuppositions without feedback/action generation step shows lower performance compared to the full \ours{} for misleading questions. 
This result demonstrates the importance of feedback/action, and also that generating presuppositions itself is not sufficient to handle misleading questions.

\section{Conclusion}
We propose \ours{}, a new approach that handles any type of question in a unified way by exploiting \textbf{presuppositions as working memory}.
\ours{} uses intermediate steps to guide the model to generate presuppositions along with useful feedback and action for itself.
Our experiments show that \ours{} boosts the performance of LLMs on different types of questions without parameter updates.
This is achieved even though the model is given no information about the type of incoming question, demonstrating that \ours{} is effective for a real-world setting and will be a significant stepping stone to future works tackling real-world QA tasks.

\section*{Limitations}
As we discussed in \autoref{subsec:ablation}, we note that misconstructed feedback and action can be a noise for LLMs, which leads them to generate incorrect answers in the answer generation step.
This phenomenon, similar to hallucination snowballing, was also pointed out by \citet{hallu_snowball}.
Future work may improve \ours{} by exploring how to prevent hallucination snowballing.

Also, our work focuses on closed-book question-answering and explores performance only on closed-book question-answering tasks.
Expanding \ours{}'s approach to tasks such as question-answering with context, text summarization, etc will be also an avenue of future work.

\bibliography{anthology}
\bibliographystyle{acl_natbib}

\appendix

\section{Designed Prompts and Few-shots}
\label{sec:appendix2-designed prompt}
\autoref{table:appendix-prompt} shows specific designed prompts and the structure of few-shots used in \autoref{main_exp}.

\begin{table*}[t]
\footnotesize	
\centering

\begin{center}
\begin{tabularx}{\textwidth}{c|X}
\toprule
Prompt &
You are a helpful assistant that analyzes the following question. Your task is to extract assumptions implicit in a given question. You must notice that considering the intention of the question will be helpful to extract a hidden assumption of the given question.\\
\midrule
Few-shot &
Question : When did the great depression began before world war 1?\par
Presuppositions : \par
(1) There was a period called the Great Depression. \par
(2) There was a conflict called World War 1. \par
(3) The Great Depression began before World War 1.\\
\bottomrule

\end{tabularx}
\end{center}

\subcaption{\scriptsize{Step 1: Presupposition Extraction. $p_A$ is added in the post-processing step.}}
\bigskip

\begin{center}
\begin{tabularx}{\textwidth}{c|X}
\toprule
Prompt &
You are a helpful assistant that provides a feedback on the question and a guideline for answering the question. You will be given a question and the assumptions that are implicit in the question. Your task is to first, provide feedback on the question based on whether it contains any false assumption and then provide a guideline for answering the question.\\
\midrule
Few-shot &
Question : When did the great depression began before world war 1?\par
Presuppositions : \par
(1) There was a period called the Great Depression. \par
(2) There was a conflict called World War 1. \par
(3) The Great Depression began before World War 1.\par
(4) There is a clear and single answer to the question. \par
Feedback : The question contains a false assumption that the Great Depression began before World War 1. \par
Action : Correct the false assumptions that the Great Depression began before World War 1 and respond based on the corrected assumption.\\
\bottomrule

\end{tabularx}
\end{center}

\subcaption{\scriptsize{Step 2: Feedback and Action generation}}
\bigskip

\begin{center}
\begin{tabularx}{\textwidth}{c|X|X}
\toprule
Method &
\ours \ &
Vanilla \\
\midrule
Prompt &
You are a helpful assistant that answers the given question. The feedback on the question and action needed for answering the question will also be given. Write a coherent and detailed answer to the question below. &
You are a helpful assistant that answers the given question. Write a coherent and detailed answer to the question below.\\
\midrule
Few-shot &
Question : When did the great depression began before world war 1?\par
Feedback : The question contains a false assumption that the Great Depression began before World War 1. \par
Action : Correct the false assumptions that the Great Depression began before World War 1 and respond based on the corrected assumption. \par
The Great Depression did not begin before World War I. The Great Depression began in 1929, which was after World War I (1914-1918). &
Question : When did the great depression began before world war 1?\par
The Great Depression did not begin before World War I. The Great Depression began in 1929, which was after World War I (1914-1918).\\
\bottomrule

\end{tabularx}
\end{center}

\subcaption{\scriptsize{Step 3: Answer generation.}}
\caption{Designed prompts and one few-shot example used in each step. \ours{} takes all three steps, while the vanilla LLM only takes the Answer Generation step. In our \autoref{main_exp}, we concat $N=6$ few shots that each have the structure shown in the table above.}
\label{table:appendix-prompt}
\end{table*}

\begin{table*}[t]
\footnotesize	
\centering

\begin{center}
\begin{tabularx}{\textwidth}{c|X}
\toprule
\textbf{Type} & \textbf{Normal Questions}\\
\midrule
\midrule
Question & 
Is the TFR1 gene dispensable for erythropoiesis?\\
\midrule
Presuppositions & 
1) There is a gene called TFR1. \par
2) It is possible for a gene to be dispensable or essential for a biological process. \par
A) There is a clear and single answer to the question.\\
\midrule
Feedback & 
\textbf{The question does not contain any false presuppositions. All the presuppositions are valid.} \\
\midrule
Action & 
Answer the question by discussing the role of the TFR1 gene in erythropoiesis and whether it is essential or dispensable for the process. Provide evidence from scientific research to support the answer.\\
\bottomrule
\end{tabularx}
\end{center}

\caption{Examples of normal question from \citep{bioasq} and the results of intermediate steps throughout our pipeline.}
\label{table:appendix-normalQ}
\end{table*}

\section{Normal Question Example}
\label{sec:appendix1-normalQ example}
Like \autoref{table:example}, we give an example of assumptions, feedback, and action on a \textit{normal question} (question without little to no imperfections). 
Numbering $A$ in the list of presuppositions stands for the additional presupposition $p_A$, which is manually added by post-processing extracted presuppositions as explained in \autoref{Presupposition Extraction}.

It can be seen in \autoref{table:appendix-normalQ} that the \textit{feedback} informs that the question does not contain any FP.
In this case, the \textit{action} functions as an elaboration of the question.

\section{Dataset} \label{sec:appendix3-dataset}
\subsection{Dataset Statistics}
\label{subsec:appendix3-datasetstat}
\para{ASQA}
ASQA includes train, development, and test split each of which contains 4,353, 948, and 1,015 questions.
For evaluation, we use the development split of ASQA because the test split of ASQA is not open publicly.

\para{(QA)$^2$}
(QA)$^2$ consists of adaptation and evaluation splits, each of which contains 32 and 570 questions, respectively.
For each split, half of them are questions that include FP, and the other half are questions that do not.
For evaluation, we use half of the evaluation split -- only questions that contain the FP.  

\para{BioASQ}
BioASQ consists of two distinct tasks (Task A, and Task B) and periodically releases datasets for each task.
We use \texttt{BioASQ9B}, which is task B released in 2021, and concatenated the 5 batches (\texttt{9B1\_golden}, \texttt{9B2\_golden}, $\cdots$, and \texttt{9B5\_golden}) in the test split for evaluation.
We regard \textit{ideal\_answer} in the dataset as a gold reference.
Following \citet{asqa}, we compare predictions against all answers in \textit{ideal\_answer} and report the maximum score. 
The concatenated test split includes 497 question-answer pairs.

\subsection{ASQA Metric}
\label{subsec:appendix-metrics}
\citet{asqa} proposed several metrics for evaluating ASQA.

\para{ROUGE-L} \citet{asqa} used a f-measure of ROUGE-L \cite{rouge} score which is a metric for evaluating generated text.

\para{Disambig-F1}
\citet{asqa} used Disambig-F1 score to evaluate the \textit{informativeness} of generated text.
The ambiguous question $q$ can be disambiguated to multiple pairs of disambiguous questions and short answers ($q_i$, $a_i$).
For each ambiguous question $q$, they feed system-generated long-form answer as a context, along with each disambiguated question ($q_i$) to a Roberta Model~\citep{roberta} that was pretrained on SQUADv2~\citep{squadv2} to predict a short answer.
Then they calculated the token-level F1 score between the predicted short answer from Roberta and the gold short answer $a_i$.
Then, Disambig-F1 score of a single ambiguous question $q$ is given by averaging the calculated F1 score of all corresponding disambiguated questions.
The final Disambig-F1 score is calculated by averaging Disambig-F1 scores of all questions in the split.

\para{Overall DR Score}
\citet{asqa} proposed a novel metric for evaluating ASQA: DR Score.
Specifically, the DR Score is given by calculating the geometric mean of Rouge-L and Disambig F1 as expressed as follows:

\(DR = \sqrt{(Disambig-F1) * Rouge}.\)

\subsection{Recall-measure Rouge-L} \label{subsec:appendix4-example_asqa}
In this part, we explain the reason for using the \textit{recall-measure} of Rouge-L instead of the f-measure with an actual example.
The ASQA dataset \citep{asqa} consists of ambiguous questions and their disambiguated versions with corresponding answers.
However, we have noticed that the disambiguated question-answer pairs in ASQA do not always cover all aspects of ambiguity inherent in the ambiguous question.
Thus, even though \ours{} generates an answer for the ambiguous question with factually correct disambiguations, if that disambiguation and answer were not provided by ASQA, using the f1-measure will penalize the factually correct answers just because they were not part of the ASQA dataset.

While we \textit{do not give higher scores} for our factually correct disambiguated answers, we thought it is fair to use a recall-measure to stop getting scores \textit{deducted} because of them.
\autoref{table:appendix-asqa-example} gives an example. 
The generated answer of \ours{} includes an explanation of "Stuck in the Middle" by Tai Verdes and "Stuck in the Middle" by Boys Like Girls which are reasonable answers to the ambiguous question. 
However, as ASQA does not include any disambiguated question-answer pairs that cover such aspects, using the f1-measure will give a even lower score compared to the prediction without those explanations.

\begin{table*}[t]
\footnotesize	
\centering
\begin{center}
\begin{tabularx}{\textwidth}{X}
\toprule
\underline{Ambiguous Question} \par Who sings the song for stuck in the middle?\\
\midrule
\underline{Disambiguated Question-Answer Pairs} \par
\textbf{Q1}) Which band sings the 1973 song "Stuck in the Middle with You"? -> \textbf{A1}) Stealers Wheel \par
\textbf{Q2}) Who sings the lead vocals in the 1973 song "Stuck in the Middle with You"? -> \textbf{A2}) Gerry Rafferty \par
\textbf{Q3}) Who sings harmony in the 1973 song "Stuck in the Middle with You"? -> \textbf{A3}) Joe Egan \par
\textbf{Q4}) Who sings the 2004 song "Stuck in the  Middle"? -> \textbf{A4}) Clea \par
\textbf{Q5}) Who sings the theme song for the TV show Stuck in the Middle? -> \textbf{A5}) Sonus
\\
\midrule
\underline{Generated answer on Ambiguous Question} \par
There are multiple songs titled "Stuck in the Middle" sung by different artists. Some of them include: \par
1. "Stuck in the Middle with You" by \textbf{Stealers Wheel}: This song was released in 1972 and was written by Gerry Rafferty and Joe Egan. It became a hit single and is often associated with the 1992 Quentin Tarantino film, Reservoir Dogs. \par
2. "Stuck in the Middle" by \textbf{Tai Verdes}: This song was released in 2020 and is part of Tai Verdes' debut album, "TV." The song gained popularity on social media platforms like TikTok. \par
3. "Stuck in the Middle" by \textbf{Boys Like Girls}: This song is from their 2006 self-titled debut album. The band is an American pop-rock group formed in 2005. \par
These are just a few examples, and there may be other songs with the same or similar titles sung by different artists. \\
\bottomrule
\end{tabularx}
\end{center}

\caption{Example from ASQA \citep{asqa} that shows why using a f1-measure for Rouge-L is inappropriate. 
The disambiguated question-answer pairs do not include all factually correct answers (Tai Verdes, Boys Like Girls), therefore penalizing generated answers even though they are more comprehensive and complete.}
\label{table:appendix-asqa-example}
\end{table*}

\section{Quality of Feedback and Action} \label{sec:appendix8-FGquality}
As discussed in \autoref{main_exp} and \ref{subsec:ablation}, poorly generated feedback and action make LLMs underperform.
~\autoref{table:appendix-FA} shows some examples of such cases.
In \autoref{table:appendix-FA} (a), it can be seen that the generated action is already itself the answer.
This makes GPT-3.5 generate the answer by merely copying the action, which does not cover the ambiguity of the question.
However, we can notice that GPT-3.5 can, in fact, discern the ambiguity of the question, by observing that the ambiguity is considered in the answer of the vanilla GPT-3.5. 
In \autoref{table:appendix-FA} (b), GPT-3.5 was misled by the generated action.
Specifically, GPT-3.5 answered that there was no fight between Muhammad Ali and Michael Dokes which is in fact not true.
The last row of (b) indicates that GPT-3.5 has knowledge about the fight between two boxers, which was not exploited due to a faulty action.

\begin{table*}[t]
\footnotesize	
\centering
\begin{center}
\begin{tabularx}{\textwidth}{c|X}
\toprule
\textbf{Type} & \textbf{Ambiguous Questions}\\
\midrule
\midrule
Question & 
Who played zordon in the original power rangers?\\
\midrule
Generated Feedback & 
The question contains a true assumption.\\
\midrule
Generated Action & 
The actor who played Zordon in the original Power Rangers series is David J. Fielding. \\
\midrule
Generated Answer & 
The actor who played Zordon in the original Power Rangers series is \textbf{David J. Fielding}.\\
\midrule
Answer by vanilla GPT-3.5 & In the original Mighty Morphin Power Rangers television series, Zordon was voiced by actor \textbf{David J. Fielding}. However, Zordon's face was portrayed by actor \textbf{Robert L. Manahan}, who was credited as "Zordon's Voice" in the show's credits.\\
\midrule
Golden Answer & In the 1993 American superhero television series Mighty Morphin Power Rangers, fictional character Zordon is initially voiced and portrayed by David Fielding, and later voiced by Robert L. Manahan. In the 1995 American superhero film Mighty Morphin Power Rangers: The Movie, Zordon is portrayed by Nicholas Bell. \\
\bottomrule
\end{tabularx}
\end{center}

\subcaption{\scriptsize{(a) Example of poorly constructed feedback and action for the ambiguous question.}}
\bigskip
\begin{center}
\begin{tabularx}{\textwidth}{c|X}
\toprule
\textbf{Type} & \textbf{Ambiguous Questions}\\
\midrule
\midrule
Question & 
Who won muhammad ali vs michael dokes?\\
\midrule
Generated Feedback & 
The question contains a false assumption that there was a fight between Muhammad Ali and Michael Dokes.\\
\midrule
Generated Action & 
Correct the false assumption that there was a fight between Muhammad Ali and Michael Dokes and respond based on the corrected assumption.\\
\midrule
Generated Answer & 
Muhammad Ali and Michael Dokes never fought against each other in a professional boxing match. Therefore, there is no winner to report.\\
\midrule
Answer by vanilla GPT-3.5 & Muhammad Ali won the fight against Michael Dokes. The fight took place on September 10, 1977, at the \"The Forum\" in Inglewood, California. Ali won the fight by a unanimous decision after 10 rounds.\\
\midrule
Golden Answer & In April 1977, Muhammad Ali fought two exhibitions of three rounds each against Jody Ballard and future World Heavyweight champion Michael Dokes, both to a No-Decision. \\
\bottomrule
\end{tabularx}
\end{center}

\subcaption{\scriptsize{(b) Example of poorly constructed feedback and action for the question that includes FP.}}
\caption{Examples of feedback and action that mislead LLMs.}
\label{table:appendix-FA}
\end{table*}

\section{Experiment on different number of few-shots} \label{sec:appendix7-diffshot}
In this part, we show additional experiments conducted using $N=4, 8$ numbers of few-shot examples (\autoref{table:appendix-4shotExp}, \autoref{table:appendix-8shotExp}).
The overall trend is consistent with \autoref{table:main_exp}, demonstrating that \ours{} is robust to different numbers of few-shot examples.

\begin{table*}[t]
\scriptsize	
\centering

\begin{center}
\begin{tabular}{c|l|ccc|ccc|ccc}
\toprule
\multirow{ 2}{*}{Model} &
\multirow{ 2}{*}{Method} &
\multicolumn{3}{c|}{(a) ASQA} & 
\multicolumn{3}{c|}{(b) (QA)$^2$} & 
\multicolumn{3}{c}{(c) BioASQ} \\

% \hline
&& Rouge-L & D-F1 & DR &
Rouge-1 & Rouge-L & BleuRT &
Rouge-1 & Rouge-L & BleuRT \\

\midrule
\midrule
GPT-4 &
Vanilla &
41.63 & 34.61 & 37.96 &
24.54 & 22.42 & 0.42 &
18.53 & 20.68 & 0.45 \\

&\ours &
\textbf{45.68} & \textbf{35.76} & \textbf{40.42} &
\textbf{27.28} & \textbf{25.33} & \textbf{0.44} &
\textbf{18.96} & \textbf{21.2} & \textbf{0.47} \\

\midrule

GPT-3.5 &
Vanilla &
43.45 & 30.74 & 36.55 &
23.99 & 22.13 & 0.41 &
\textbf{19.94} & \textbf{22.35} & 0.46 \\

&\ours &
41.22 & 28.36 & 34.19 &
26.78 & 24.93 & 0.43 &
17.31 & 19.79 & 0.46 \\

&~ - ~ w. GPT-4 $(F,A)$&
\textbf{45.83} & \textbf{30.79} & \textbf{37.57} &
\textbf{27.13} & \textbf{25.42} & \textbf{0.44} &
18.96 & 21.82 & \textbf{0.47} \\
\bottomrule
\end{tabular}
\end{center}

\caption{Result of \ours{} and vanilla GPT-4 and GPT-3.5 using $N=4$ few-shot examples. The last row (w. GPT-4 $(F,A)$) refers to the system that generates answers on GPT-3.5 using feedback and action generated by GPT-4.}
\label{table:appendix-4shotExp}
\end{table*}

\begin{table*}[t]
\scriptsize	
\centering

\begin{center}
\begin{tabular}{c|l|ccc|ccc|ccc}
\toprule
\multirow{ 2}{*}{Model} &
\multirow{ 2}{*}{Method} &
\multicolumn{3}{c|}{(a) ASQA} & 
\multicolumn{3}{c|}{(b) (QA)$^2$} & 
\multicolumn{3}{c}{(c) BioASQ} \\

% \hline
&& Rouge-L & D-F1 & DR &
Rouge-1 & Rouge-L & BleuRT &
Rouge-1 & Rouge-L & BleuRT \\

\midrule
\midrule
GPT-4 &
Vanilla &
\textbf{42.51} & 34.98 & 38.56 &
25.72 & 23.68 & 0.43 &
18.82 & 20.95 & 0.45 \\

&\ours &
42.07 & \textbf{35.82} & \textbf{38.82} &
\textbf{33.53} & \textbf{31.68} & \textbf{0.46} &
\textbf{21.55} & \textbf{24.15} & \textbf{0.46} \\

\midrule

GPT-3.5 &
Vanilla &
41.94 & 30.28 & 35.64 &
25.46 & 23.21 & 0.42 &
20.68 & 23.26 & 0.46 \\

&\ours &
40.88 & 28.73 & 34.27 &
28.60 & 25.67 & 0.44 &
19.61 & 22.10 & 0.46 \\

&~ - ~ w. GPT-4 $(F,A)$ &
\textbf{45.39} & \textbf{31.38} & \textbf{37.74} &
\textbf{30.38} & \textbf{28.11} & \textbf{0.45} &
\textbf{21.65} & \textbf{24.35} & 0.46 \\
\bottomrule
\end{tabular}
\end{center}

\caption{Result of \ours{} and vanilla GPT-4 and GPT-3.5 using $N=8$ few-shot examples. The last row (w. GPT-4 $(F,A)$) refers to the system that generates answers on GPT-3.5 using feedback and action generated by GPT-4.}
\label{table:appendix-8shotExp}
\end{table*}

\section{Human Evaluation Prompt} \label{sec:appendix8-human_eval_prompt}
In this part, we release the prompt that are given to human evaluators in \autoref{table:appendix-humanEvalPrompt}.

\begin{table*}[t]
\scriptsize	
\centering
\begin{center}
\begin{tabularx}{\textwidth}{X}
\toprule
In this task, you will be shown one question and its golden answer, and two generated answers from different systems.\par 
Your goal is to evaluate which answer is better for each of the three criteria (Overall Impression, Completeness, Correctness) by referring to the given golden answer.\par 
For the Overall Impression criterion, your job is to select an answer (or mark it as a tie) that gives you more satisfaction. Fluency, correctness, consistency, sufficiency of information, or even formatting can be considered as a factor.\par  
For the Completeness criterion, your job is to select an answer (or mark it as a tie) considering whether the answer provides enough information. You can refer to the given golden answer for the evaluation.\par 
For the Correctness criterion, your job is to select an answer (or mark it as a tie) considering whether the answer does not include any hallucination (factually wrong information).\par 
You can refer to the given golden answer for the evaluation. \\
\bottomrule
\end{tabularx}
\end{center}

\caption{Prompt given to human evaluator}
\label{table:appendix-humanEvalPrompt}
\end{table*}

\end{document}